\documentclass[10pt,twocolumn,letterpaper]{article}

\usepackage{wacv}              

\usepackage{graphicx}
\usepackage{amsmath}
\usepackage{amssymb}
\usepackage{multirow}

\usepackage{booktabs}
\usepackage{upgreek}
\usepackage{enumitem}
\usepackage{caption}
\usepackage[boxed, ruled, vlined, linesnumbered]{algorithm2e}
\usepackage[aboveskip=0pt, belowskip=0pt]{caption}
\usepackage{amsthm}
\theoremstyle{definition}
\newtheorem{definition}{Definition}[section]
\usepackage{bm}

\newcommand{\reg}[0]{{\rm reg}}

\newcommand{\Dist}[0]{{\rm Dist}}

\usepackage[pagebackref,breaklinks,colorlinks]{hyperref}

\usepackage[capitalize]{cleveref}
\crefname{section}{Sec.}{Secs.}
\Crefname{section}{Section}{Sections}
\Crefname{table}{Table}{Tables}
\crefname{table}{Tab.}{Tabs.}

\def\wacvPaperID{1457} 
\def\confName{WACV}
\def\confYear{2025}

\begin{document}

\title{Invariant Shape Representation Learning For Image Classification}

\author{Tonmoy Hossain\textsuperscript{1*}, Jing Ma\textsuperscript{1\dag}, Jundong Li\textsuperscript{1,2*}, Miaomiao Zhang\textsuperscript{1,2*}\\
Computer Science\textsuperscript{1}, Electrical and Computer Engineering\textsuperscript{2} \\
University of Virginia\textsuperscript{*}, Case Western Reserve University\textsuperscript{\dag}\\
{\tt\small \{tonmoy,jl6qk,mz8rr\}@virginia.edu\textsuperscript{*}, jing.ma5@case.edu\textsuperscript{\dag}}
}
\maketitle

\begin{abstract}
   Geometric shape features have been widely used as strong predictors for image classification. Nevertheless, most existing classifiers such as deep neural networks (DNNs) directly leverage the statistical correlations between these shape features and target variables. However, these correlations can often be spurious and unstable across different environments (e.g., in different age groups, certain types of brain changes have unstable relations with neurodegenerative disease); hence leading to biased or inaccurate predictions. In this paper, we introduce a novel framework that for the first time develops invariant shape representation learning (ISRL) to further strengthen the robustness of image classifiers. In contrast to existing approaches that mainly derive features in the image space, our model ISRL is designed to jointly capture invariant features in latent shape spaces parameterized by deformable transformations. To achieve this goal, we develop a new learning paradigm based on invariant risk minimization (IRM) to learn invariant representations of image and shape features across multiple training distributions/environments. By embedding the features that are invariant with regard to target variables in different environments, our model consistently offers more accurate predictions. We validate our method by performing classification tasks on both simulated 2D images, real 3D brain and cine cardiovascular magnetic resonance images (MRIs). Our code is publicly available at \href{https://github.com/tonmoy-hossain/ISRL}{\tt https://github.com/tonmoy-hossain/ISRL}.
   
\end{abstract}

\section{Introduction}
\label{sec:intro}

Deformable shape has a long history of aiding image analysis, as it plays an important role in the processing of human visual information~\cite{zhang2016statistical,hong2017fast,reaungamornrat2017deformable,jayakumar2023sadir}. Recent works on shape-based deep networks have demonstrated the robustness of shape to variations in image intensity and texture (e.g., noisy or corrupted data) for image analysis tasks, such as classification~\cite{geirhos2018imagenet,hossain2019brain,wang2022geo}. Despite achieving improved performance in classification accuracy, the aforementioned models were designed to leverage any statistical correlations between the geometric shape features and target image labels. However, these features can contain both ``invariant'' features that have stable relationships with the labels under any circumstances, and ``spurious'' features that have varying relationships with labels across different contexts or environments (e.g., a confounding factor can often lead to spurious correlations).


Existing shape-based image classification networks are often incapable of distinguishing the invariant shape features from the spurious ones~\cite{geirhos2018imagenet,wang2022geo}; hence are easily fooled by illusory patterns and perform unstably in different environments. For example, anatomical brain changes derived from images have been used as predictive features to distinguish healthy subjects from neurodegenerative diseases~\cite{wen2020convolutional,wang2022geo}. However, in different environments such as age groups, if not properly handled (e.g., when the average age of the disease group is significantly older than the healthy controls), a predictor (classifying healthy/disease) can be potentially confounded~\cite{zhao2020training,zare2022removal}, rather than capturing the invariant features caused by Diseases. Teasing apart spurious features related to age factor is critical for DNNs to provide robust predictions of brain diseases. 

A few research groups have initial attempts to capture more reliable information instead of spurious correlations to facilitate the robustness of image classification under different scenarios, mainly including confounding biases~\cite{yue2020interventional,zhao2020training} and weak/noisy supervision~\cite{castro2020causality,zhang2020causal,yang2023treatment}. Invariant learning is one of the subsets of this arena that recently attracted significant attention \cite{arjovsky2019invariant,ahuja2020invariant,chang2020invariant}. The goal of invariant learning is to identify and capture the underlying factors that have a stable relationship with the label across different environments, while disregarding factors with unstable spurious correlations to the label. More intuitively, it aims to capture the real “cause” of the label to enable more robust and accurate predictions, improving outcomes even for unseen environments, which results in achieving promising performance of extracting invariant features in the image space
~\cite{arjovsky2019invariant,ahuja2020invariant,nam2020learning,mitrovic2020representation,lu2021invariant,wang2022out}. Another line of work investigated domain generalization approaches to deal with possible changes across known input distributions by exploiting unsupervised adaptation~\cite{sun2016return,sun2016deep}, domain calibration~\cite{wald2021calibration,gong2021confidence}, data augmentation~\cite{li2021simple,yao2022improving,hossain2023mgaug}, or adversarial learning~\cite{ganin2016domain,yang2021adversarial}. However, none of the existing studies have investigated invariant learning methods in an integrated shape space. This limits their ability to fully utilize geometric features that have been proven to be important and robust in image classification tasks.

In this paper, we introduce a novel method that for the first time develops a joint learning of {\em invariant shape representations} for improved performance of image classifiers. In contrast to previous approaches that are limited to learning invariant features solely in the image spaces~\cite{zhao2020training,terziyan2023causality,yang2023treatment}, our algorithm has the main advantages of
\begin{enumerate}[label=(\roman*)]
\item Developing a new learning paradigm based on invariant risk minimization (IRM)~\cite{ahuja2020invariant} to learn integrated image and shape representations that are invariant across multiple training distributions/environments. 
\item Improving the efficiency and adaptability of image classifiers when tested with unseen data/environments, by leveraging learned invariant features with maximally eliminated spurious correlations.
\item Opening new avenues for causal representation learning by leveraging invariant image and shape features associated with target labels.


\end{enumerate}
We validate the effectiveness of ISRL on 2D simulated data~\cite{jongejan2016quick}, 3D real brain MRIs~\cite{jack2008alzheimer}, and 3D cardiac MRI videos~\cite{wang2022ai}. Experimental results show that our method outperforms the state-of-the-arts by significantly improved robustness to shifted environments with consistently higher classification accuracy.

\section{Background: Deformation-based Shape Representations}
The literature has studied various representations of geometric shapes, including landmarks~\cite{cootes1995active}, binary segmentations~\cite{brechbuhler1995parametrization}, and medial axes~\cite{pizer1999segmentation}. These aforementioned techniques often ignore objects' interior structures; hence do not capture the intricacies of complex objects in images. In contrast, deformation-based shape representations (based on elastic deformations or fluid flows) focus on highly detailed shape information from images~\cite{vaillant2004statistics,lorenzen2005unbiased,bone2018learning}. This paper will feature deformation-based shape representations that offer more flexibility in describing shape changes and variability of complex structures. However, our developed methodology can be easily adapted to other representations, including those characterized by landmarks, binary segmentations, curves, and surfaces. With the underlying assumption that objects in many generic classes can be described as deformed versions of an ideal template, descriptors in this class arise naturally by transforming/deforming the template to an input image~\cite{lorenzen2005unbiased}. The resulting transformation is then considered as a shape that reflects geometric changes.

Given a number of $N$ images, $\{I_1,\cdots,I_N\}$, the problem of template-based image registration is to estimate the deformation fields, $\{\phi_1,\cdots, \phi_N\}$, between a template image $I$ and each individual image via minimizing the energy
\setlength{\abovedisplayskip}{0pt}
\begin{align}
\label{eq:lddmm}
E(I, \phi_n(t)) &=  \sum_{n=1}^{N} \frac{1}{\sigma^2} \text{Dist} [I \circ \phi^{-1}_n(v_n(t)), I_n] \nonumber \\  &+ \int_0^1 (L v_n(t), v_n(t)) \, dt,
\end{align}
subject to $d\phi_n(t) / dt = v_n(t) \circ \phi_n(t)$. Here, $\sigma^2$ is a noise variance and $\circ$ denotes an interpolation operator that deforms image $I$ with an estimated transformation $\phi_n$. The $\Dist[\cdot, \cdot]$ is a distance function that quantifies the dissimilarity between images, i.e., sum-of-squared differences~\cite{beg2005computing}. 
Here, $L: V\rightarrow V^{*}$ is a symmetric, positive-definite differential operator that maps a tangent vector $ v_n(t)\in V$ into its dual space as 
a momentum vector $m_n(t) \in V^*$. We write $m_n(t) = L v_n(t)$, or $v_n(t) = K m_n(t)$, with $K$ being an inverse operator of $L$. The notation $(\cdot, \cdot)$ denotes the pairing of a momentum vector with a tangent vector, which is similar to an inner product. In this paper, we use a metric of the form $L = -\alpha \Delta + \mathbb{I}$, in which $\Delta$ is the discrete Laplacian operator, $\alpha$ is a positive regularity parameter, and $\mathbb{I}$ denotes an identity matrix.  

The {\em geodesic shooting} algorithm~\cite{vialard2012diffeomorphic} states that the minimum of Eq.~\eqref{eq:lddmm} is uniquely determined by solving an Euler-Poincar\'{e} differential equation (EPDiff)~\cite{arnold1966geometrie,miller2006geodesic} with a given initial condition of velocity field $v_n(0)$,

\begin{figure*}[t]
    \centering
    \includegraphics[width=\textwidth, trim = 0cm 0cm 0cm 0cm]{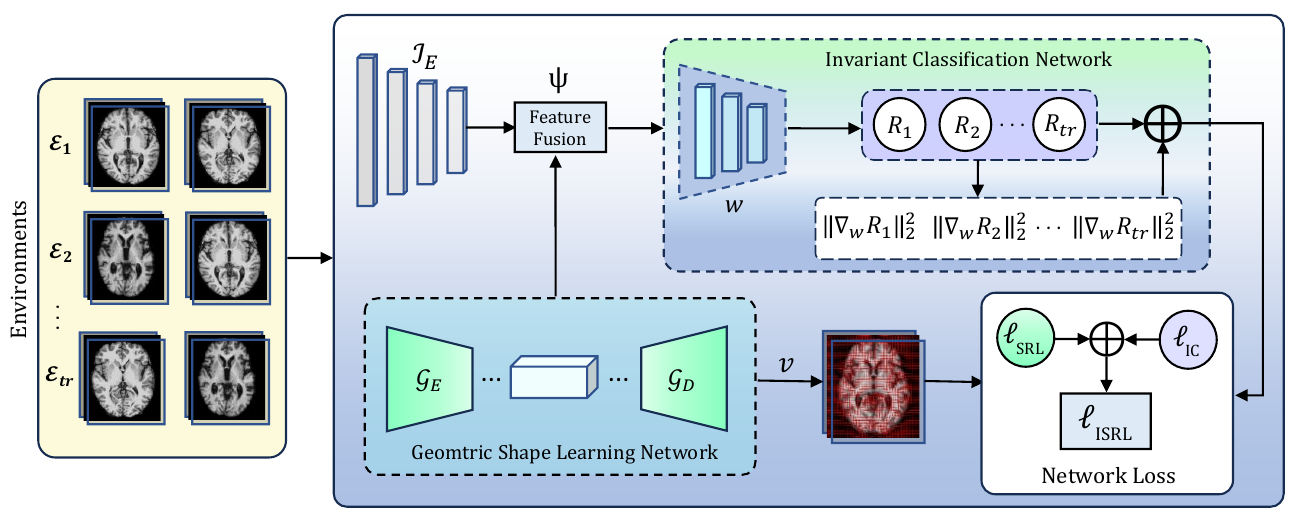} 
    \caption{An overview of our proposed network architecture of ISRL. The geometric shape learning ($\mathcal{G}_E , \mathcal{G}_D$) and image network ($\mathcal{I}_E$) is taking images from different environments \{$\mathcal{E}_1,\cdots, \mathcal{E}_{tr}$\}. ISRL combined features from latent spaces passing it to the classifier $w$. To learn invariant features, we combine geometric shape learning loss with environment-wise risk ($R_{tr}$) along with their gradient ($||\nabla_w R_{tr}||_2^2$). }  
    \label{fig:model}
\end{figure*}

\setlength{\abovedisplayskip}{0pt}

\begin{align}
    \frac{\partial v_n(t)}{\partial t} &= - K [(D v_n(t))^T \cdot m_n(t) + D m_n(t) \cdot v_n(t) \nonumber \\ &+ m_n(t) \cdot \operatorname{div} v_n(t) ],
    \label{eq:epdiff}
\end{align}
where $D$ denotes a Jacobian matrix, $\operatorname{div}$ is the divergence, and $\cdot$ represents element-wise matrix multiplication. 

We are now able to equivalently minimize the energy function in Eq.~\eqref{eq:lddmm} as

\setlength{\abovedisplayskip}{0pt}

\begin{align}
   E(I, v_n(0)) &=  \sum_{n=1}^{N} \frac{1}{\sigma^2} \text{Dist} [I \circ \phi^{-1}_n(v_n(t)), I_n] \nonumber \\ &+ (L v_n(0), v_n(0)), \, \,  \text{s.t. Eq.~\eqref{eq:epdiff}.}   
\label{eq:flddmm}
\end{align}
For notation simplicity, we will drop the time index in the following sections. 

\section{Our Model}
This section presents a novel algorithm, ISRL, that for the first time develops a joint learning of invariant shape representations for improved image classifiers.

\noindent {\bf Problem setup.} Given a number of $J$ image classes, there exists a number of $N_j, j \in \{1, . . . , J\}$ images in each class. With a group of training images $\{I_{nj}\}_{n=1,j=1}^{N_j,J}$ and their associated class labels $\{y_{nj}\}$ collected from multiple training environments $\{e\} \in \mathcal{E}_{tr}$, we define the training data as $X = \{(I_{nj}^{e}, y_{nj}^{e})\}_{e=1}^{\mathcal{E}_{tr}}$. Note that each image $I_{nj}$ is considered as a deformable variant of a template $T$, where $T \in N_j$. The proposed framework of ISRL consists of two modules: (i) an unsupervised learning of geometric deformations via a template-based registration network~\cite{wang2022geo}, and (ii) an invariant classification network that takes fused features from latent spaces of image intensities and geometric shapes. An overview of our network architecture is shown in Fig.~\ref{fig:model}. 


\subsection{Geometric Shape Representation Learning Network} 
Let $\Uptheta = (\mathcal{G}_{E}, \mathcal{G}_{D})$ be the parameters of an encoder-decoder in our geometric shape learning network. The shape representations of images within each class (a.k.a., initial velocity fields $v_{nj}(\Uptheta)$) will be learned by minimizing the network loss function as
\begin{align}
\ell_{\text{SRL}}(\Uptheta) &= \sum_{n=1}^{N_{j}}\sum_{j=1}^{J} [\frac{1}{\sigma^{2}} \|T \circ \phi^{-1}_{nj}(v_{nj}(\Uptheta)) - I_{nj}\|_{2}^{2} + \reg(\Uptheta)  \nonumber\\
&+ (\mathcal{L}_{j} v_{nj}(\Uptheta), v_{nj}(\Uptheta))], \,\, \text{s.t. Eq.~\eqref{eq:epdiff}},
\label{eq:isr}
\end{align}
where $\reg(\cdot)$ denotes a regularization term on network parameters. Note that this shape representation learning network takes into account images within each class across all environments, i.e., $I_{nj} \triangleq \{I_{nj}^{e}\}_{e=1}^{\mathcal{E}_{tr}}$. For notation simplicity, we omit the environment variable $e$.  

The template $T$ can be either treated as network parameters to estimate~\cite{hinkle2018diffeomorphic,wang2022geo}, or pre-selected references~\cite{kim2015improved,ye2018atlas}. In this experiment, we use a pre-selected image as a reference template. As for the shape learning network backbone, we adopt a commonly used UNet architecture~\cite{ronneberger2015u} in this paper. However, other networks such as UNet++ \cite{zhou2018unet++} and TransUNet~\cite{chen2021transunet} can be easily applied.

\subsection{Invariant Classification Network} 
Our newly designed classification network is trained to learn the unobserved {\em invariant features from an integrated image and deformation shape spaces}, aiming to achieve an optimal performance.
Inspired by recent works~\cite{ahuja2020invariant,arjovsky2019invariant}, we develop a mechanism of invariant representation learning that captures both geometric shape and image features robust to data distribution shifts across multiple environments. Such features may serve as indicators or proxies for the underlying invariant relationships to determine the labels of image classes. Meanwhile, the biases introduced by spurious factors, will be effectively mitigated. 

Let $\mathcal{I}_E$ be the parameters of an encoder that extracts features from image spaces. We first employ a feature fusion module that concatenates the representation of geometric shape, $\mathbf{v}$, and image, $\mathbf{I}$, in a latent space $\mathcal{H}$~\cite{wang2022geo}. Note that $\mathcal{I}_E$ can be a wide variety of feature extractors, including but not limited to ResNet~\cite{he2016deep}, VGGNet~\cite{simonyan2014very}, ViT~\cite{dosovitskiy2020image}, or other state-of-the-art network architectures for image classification tasks. An invariant classifier that maps the latent features in $\mathcal{H}$ to predicted labels, $\mathbf{y}$, is defined as follows.  

\begin{definition}
Consider an integrated shape and image representation $\uppsi(\mathcal{G}_{E}, \mathcal{I}_E):\mathbf{v} \times \mathbf{I}  \rightarrow \mathcal{H}$. A classifier $w(\uppsi)$ is invariant across environments $\{e\}$, when there exists a model $w:\mathcal{H} \rightarrow \mathbf{y}$  simultaneously optimal for all environments. That is to say, we have $w \in argmin \sum_{e\in \mathcal{E}} R_e(\uppsi)$, where $R_e(\uppsi) := \mathbb{E}_{\mathbf{I}_e, \mathbf{v}_e, \mathbf{y}_e} [l(\uppsi(\mathbf{I}_e, \mathbf{v}_e, \mathbf{y}_e)]$ is the empirical risk under environment $e$.
\end{definition}

Following a practical implementation of IRM~\cite{arjovsky2019invariant}, we are now ready to define a loss function of our invariant classification network as a combination of the sum of empirical risk minimization with invariant constraints of the predictor $w$. With $\mathcal{E}_{tr}$ denoting a set of all training environments, we formulate the loss of our invariant classification network as

\setlength{\abovedisplayskip}{0pt}

\begin{equation}
\ell_{\text{IC}}(\uppsi) = \underset{\uppsi}{min}\sum_{e\in \mathcal{E}_{tr}} R_e(\uppsi) + \lambda\left\|\bigtriangledown_{w|w=1.0} R_e(w\cdot\uppsi)\right\|_{2}^2,
    \label{eq:irm}
\end{equation}
where $w = 1.0$ is now a scalar (or a fixed dummy classifier) and $\uppsi$ becomes the entire invariant predictor. The parameter $\lambda$ is a penalty weight to control the invariance of the predictor $\uppsi$ across different training environments. Minimizing the loss above encourages the learned feature representations to discard spurious features but embed invariant factors for a more robust classification performance.

In this paper, we employ a cross-entropy loss as the classification loss $R_e$ in each environment, i.e.,
\begin{align}
R_e(\uppsi(\mathcal{G}_{E}, \mathcal{I}_E)) &= \tau \sum_{n=1}^{N_{j}}\sum_{j=1}^{J} -y_{nj} \cdot \log \hat{y}_{nj}(\uppsi(\mathcal{G}_{E}, \mathcal{I}_E)) \nonumber \\ &+ \reg(\uppsi(\mathcal{G}_{E}, \mathcal{I}_E)),
\label{eq:clf}
\end{align}
where $\tau$ is a weighting parameter and $\hat{y}_{nj}$ denotes a predicted label of the $n$-th subject in class $j$. 

\subsection{Joint Learning Network Loss and Optimization.} 

The loss function of our joint learning framework, ISRL, includes the loss from both geometric learning (Eq.~\eqref{eq:isr}) and invariant classification networks (Eq.~\eqref{eq:irm}). Defining $\beta$ as the weighting parameter, we are now ready to write the joint network loss as $\ell_{\text{ISRL}} = \ell_{\text{SRL}}(\Uptheta(\mathcal{G}_E,\mathcal{G}_D)) + \beta \ell_{\text{IC}}(\uppsi(\mathcal{G}_E,\mathcal{I}_E))$. We employ an alternative optimization scheme~\cite{nocedal1999numerical} to minimize the total loss. More specifically, we jointly optimize all network parameters by alternating between the training of the geometric shape learning and invariant classification network, making it an end-to-end learning \cite{chaitanya2021semi, chen2022enhancing}.  

The training inference of ISRL is summarized in Alg.~\ref{alg:altopt}.

\begin{algorithm}[t]
\SetAlgoLined
\SetArgSty{textnormal}
\SetKwInOut{Input}{Input}
\SetKwInOut{Output}{Output}
  \Input{A group of input images ${\mathbf{I}}$ with their associated class labels ${\mathbf{y}}$, number of environments $tr$, and convergence threshold $\epsilon$.} 
  \Output{Predicted initial velocity fields $\mathbf{v}$ and class labels $\hat{\mathbf{y}}$.}

\Repeat{\text{convergence,} $|\Updelta \ell_{\text{ISRL}}| < \epsilon$}{
\For{$e \in \forall \mathcal{E}_{tr}$}
   {

   \tcc{Train our geometric shape learning network}
   
   Optimize the geometric shape learning loss in Eq.~\eqref{eq:isr};
   
   Output the initial velocity fields $\mathbf{v}$ and within group template image $\{T_j\}$ (if treated as network parameters rather than pre-selected images) for all classes.
   
   \tcc{Train our invariant classifier}

   Output the predicted class labels $\hat{\mathbf{y}}$.
   
   Optimize the invariant classifier loss in Eq.~\eqref{eq:irm} with integrated features from shape and image spaces;
   }
}
\caption{Joint optimization of ISRL model.}
\label{alg:altopt}
\end{algorithm}

\section{Experimental Evaluation}

\begin{figure}[t]
    \centering
    \includegraphics[width=1.0\linewidth, trim = 0cm 0cm 0cm 0cm]{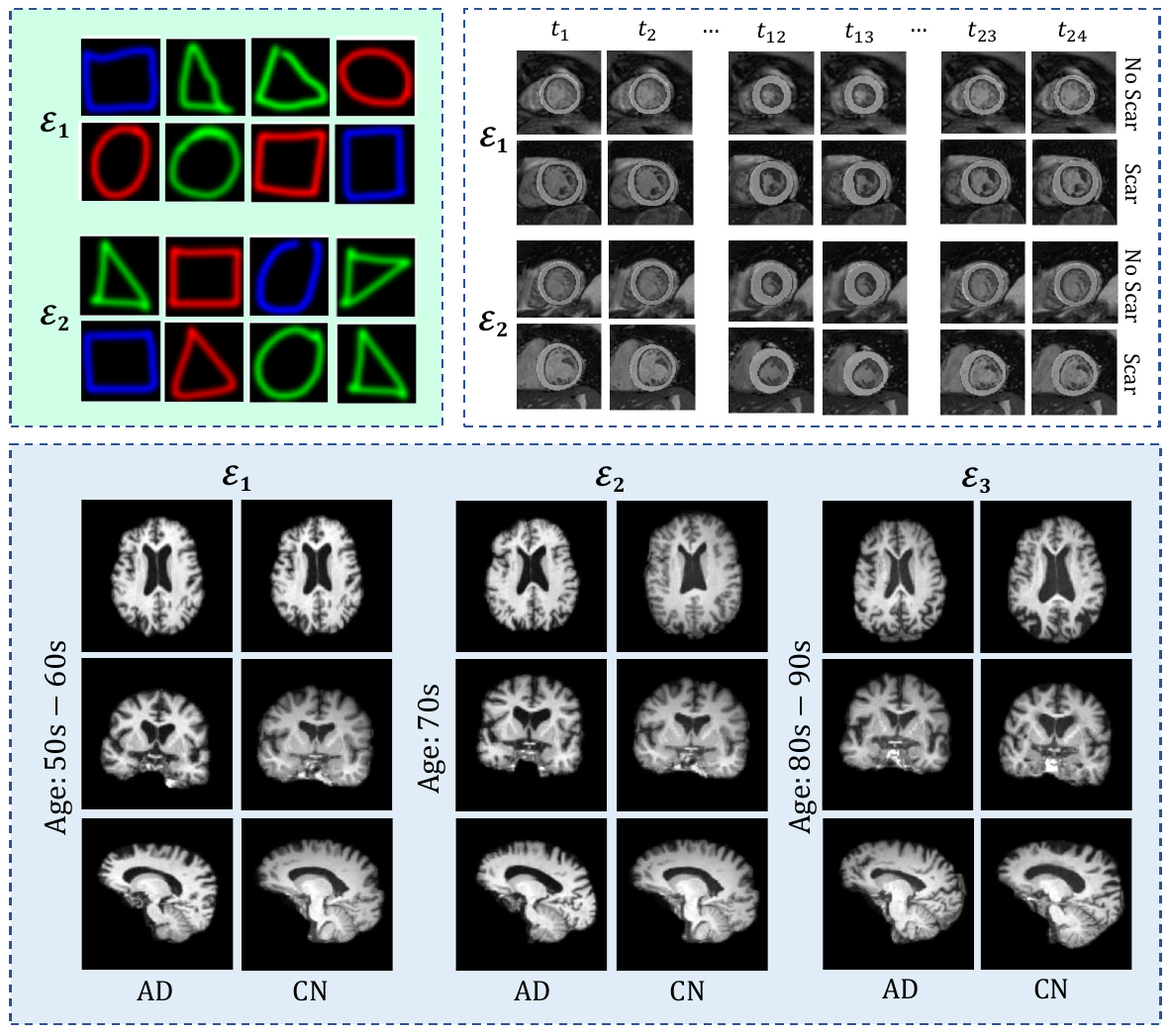} 
    \caption{Top to bottom: examples of 2D simulated data vs. 3D brain MRIs vs. 3D video sequence of cardiac MRIs. The training environments of each dataset (color vs. age vs. patient history of congestive heart failure).}
    \label{fig:datasets}
\end{figure}

We demonstrate the effectiveness of our proposed model on a diverse set of deformable image datasets, including 2D simulated data, 3D real brain MRIs, and 3D video sequences of cardiac MRIs. Examples of all datasets in multiple different training environments (color vs. age vs. patient history of congestive heart failure) are shown in Fig.~\ref{fig:datasets}.

\noindent{\textbf{2D simulated data.}} We first randomly choose $3000$ 2D images with three different classes of circles, squares, and triangles ($1000$ images per class) from the Google Quickdraw dataset~\cite{jongejan2016quick}. All images underwent affine transformation and intensity normalization with the size of $224\times224$. 

Analogous to~\cite{arjovsky2019invariant}, we first assign a random color (red, green, or blue) to each image, introducing a spurious correlation between the color information and class labels. Such a correlation is artificially established to make color more predictive of the label than the actual drawings. As a result, current algorithms focused solely on minimizing training errors, such as ERM (an image classifier extracting features from image space by minimizing the empirical classification training loss) \cite{vapnik1991principles}, will tend to exploit the color. In this scenario, we expect our proposed algorithm, ISRL, is able to (i) observe the strength of the correlation between color and label varies across multiple training environments; and (ii) leverage the geometric shape information derived from images to eliminate color as a predictive feature; leading to enhanced generalization performance.

Following \cite{arjovsky2019invariant}, we then synthesize a colored Google Quickdraw dataset with three environments, including two training and one testing by performing the following steps: i) assign a preliminary label $\Tilde{y}$: $\Tilde{y} = 0$ for circle, $\Tilde{y} = 1$ for square, and $\Tilde{y} = 2$ for triangle; ii) obtain the final label $y$ by flipping $\Tilde{y}$ with probability $p^l$ (i.e., $0.25$, $0.4$, and $0.5$ in our experimental settings); iii) sample the color id $z$ by flipping $y$ with different probabilities $p^e$ in each environment. In this paper, we set $p^e = 0.2, 0.1, 0.9$ for the two training and one testing environment, respectively. We then color the image red if $z=0$, green if $z=1$, and blue if $z=2$. We split such a synthesized dataset into $70\%$ for training, $15\%$ for validation, and $15\%$ for testing.

\noindent {\bf 3D brain MRI.} For Alzheimer's Disease (AD) classification, we include $690$ public T1-weighted brain MRIs from the AD Neuroimaging Initiative (ADNI)~\cite{jack2008alzheimer}. All subjects ranged in age from $50$ to $100$, with $345$ images from patients affected by AD and the others from cognitively normal (CN). All MRIs were preprocessed to be the size of $104\times128\times120$, $1mm^{3}$ isotropic voxels, and underwent skull-stripping, intensity min-max normalization, bias-field correction, and affine registration~\cite{reuter2012within}. 

This experiment focuses on treating age as an environment, which is motivated by the well-established understanding that the brain changes caused by different ages are often confused with the brain changes related to AD ~\cite{raji2009age,zhao2020training,rasal2022deep}. That is, at different ages, certain types of brain changes have varying correlations with AD.
In this setting, we expect the model to learn invariant features of brain shape changes that are actual diagnostic criteria for AD. The spurious features are brain shape changes affected by age. In this context, we set the training environments, $\mathcal{E}_{tr}$, of our model into three categories representing individuals in their $50$s to $60$s, $70$s, and $80$s to $90$s. Each age group or environment is characterized by a uniform distribution and includes a total of $150$ images. We split the data into $65\%$, $25\%$, and $10\%$ as training, testing, and validation.

\noindent {\textbf{3D cardiac MRI videos.}} We include $510$ cine cardiac MRI video sequences with manually delineated left ventricular myocardium segmentation maps collected from $125$ subjects~\cite{wang2022ai}. Approximately $40\%$ of these subjects exhibit heart motion abnormalities caused by myocardial scars. All videos were preprocessed to be the size of $224\times224\times24$, $1mm^{3}$ isotropic voxels, where $24$ represents the number of time frames that cover a full cardiac cycle. 

In this experiment, we perform patient scar classification (scar vs. non-scar) from the 3D video sequence of myocardium contour segmentation. We treat congestive heart failure (CHF) as an environment, given evidence indicating that individuals with CHF have an increased risk of heart motion abnormalities~\cite{podrid1992ventricular,kjekshus1990arrhythmias}. We expect to learn invariant shape features of myocardial motion changes that are actual indicators for the scar, while the spurious features are motions led by CHF. In this context, the two training environments $(\mathcal{E}_{1}, \mathcal{E}_{2})$, are subjects labeled as having CHF or not. Out of the $510$ videos (approximately equal number of scar/non-scar), we found $146$ videos have CHF ($\mathcal{E}_{1}$, $48$/$98$ as scar/non-scar) and $364$ not ($\mathcal{E}_{2}$, $210$/$154$ as scar/non-scar). We split the data into $70\%$, $15\%$, and $15\%$ for training, validation, and testing, respectively.
\captionsetup[figure]{belowskip=0pt}

\subsection{Experiment Settings}
\label{sec:exp_base}

We evaluate ISRL from two perspectives: (i) the benefits of incorporating the learning of geometric shape features for classification and (ii) the effectiveness of our developed invariant representation learning. We compare our model's performance with two categories of baseline algorithms: a typical image classifier extracting features solely from image space based on ERM~\cite{vapnik1991principles} and a recent work that learns invariant classification features solely from the image space based on IRM~\cite{arjovsky2019invariant}. 

We also compare ISRL with four state-of-the-art domain generalization models, including CORAL~\cite{sun2016deep}, DANN~\cite{ganin2016domain}, IWDAN~\cite{tachet2020domain}, and CLOvE~\cite{wald2021calibration} on 3D Brain and Cardiac MRI experiments. Please note that none of these algorithms explicitly model shape feature learning.

\noindent {\bf Classification evaluation.}  We compare all methods with four classification network backbones on both simulated 2D and real 3D brain MRI datasets, including a custom four-block convolutional neural network (CNN), AlexNet~\cite{krizhevsky2017imagenet}, ResNet18~\cite{he2016deep}, and VGGNet11~\cite{simonyan2014very}. For the custom CNN, each block comprises a convolutional layer with a kernel size of $5\times5$, a max-pooling layer with a size of $3\times3$, and batch normalization. We examine the micro-averaged accuracy (Acc.), precision (Prec.), and F1-score (F1-sc.) for the 2D simulated and 3D brain experiments. For all experiments on the 3D cardiac MRI videos, we use vision transformer (ViT)~\cite{dosovitskiy2020image} and Video-ViT (ViViT)~\cite{arnab2021vivit} as the network backbone. We evaluate the accuracy and F1-score for classification performance.

To investigate the robustness of ISRL on 2D simulated dataset, we manipulate the images by increasing the number of flipped labels $p^e$ (i.e., from 10\% to 50\%) \cite{arjovsky2019invariant} and compare the accuracies of the baselines on all network backbones. Besides, we cover an ablation study of ISRL measuring the effectiveness of shape information and invariant learning scheme on the network's prediction.

\noindent {\bf Evaluation on joint learning.} To evaluate the effectiveness of jointly learning the invariant shape representations along with the classification task, we conduct an analysis by comparing ISRL with a disjoint two-step training approach. In the latter, our geometric shape learning network is employed as an initial preprocessing stage to extract geometric shape features. These extracted features are then integrated with image features, which are the inputs to the invariant classification network. We run validation on both 2D simulated data and 3D brain MRIs.

\noindent {\bf Ablation on invariant shape learning module.} We perform an ablation study comparing our invariant shape learning model ISRL against non-invariant ERM (with and without shape features) and invariant IRM (omitting shape features) on all datasets.

\noindent {\bf Parameter setting.} We set the number of time steps for Euler integration in EPDiff (Eq.~\eqref{eq:epdiff}) as $10$, and the noise variance $\sigma = 0.02$. Besides, we set the parameter $\alpha=3$ for the operator $\mathcal{L}$, and the batch size as $32$ and $8$ for 2D and 3D experiments, respectively. For network training, we utilize the cosine annealing learning rate scheduler that starts with a learning rate of $\eta=1\text{e-}3$. 
We perform a cross-validation with increased penalty weights on all datasets to determine an optimal parameter $\lambda$ that governs the strength of the invariance penalty in the network loss (in Eq.~\eqref{eq:clf}) of our ISRL model, and the baseline algorithm IRM. All models are trained by the Adam optimizer with the best validation performance until convergence.  

\subsection{Results}
Tab.~\ref{table:clf_SOTA_comp} reports the micro-averaged classification performance on 2D colored Google Quickdraw (with $p^l=0.25$, meaning $25\%$ flipped labels) for all backbones. Our proposed model consistently outperforms the baselines across all backbones. More specifically, classifiers that incorporate shape features, (such as our ISRL), demonstrate notably enhanced accuracy compared to classifiers that rely solely on image intensities (IRM / ERM). Furthermore, as expected, classifiers utilizing invariant learning (our ISRL or IRM) are more robust to spurious correlations introduced by colors, resulting in much higher classification accuracy when compared to classifiers without invariant learning (ERM).
\begin{table}[htbp]
\centering
\caption{A comparison of classification performance on simulated 2D data over all models with different network backbones.}
\begin{tabular}{lcccc}
\hline
\textit{Model} & \textit{Network}         & \textit{Acc}   & \textit{Prec}  & \textit{F1-sc} \\ \hline
ERM            & \multirow{3}{*}{CNN}     & 48.67          & 48.52          & 48.17          \\  
IRM            &                          & 68.67          & 68.86          & 68.61          \\ 
\textbf{ISRL} &                          & \textbf{78.89} & \textbf{79.64} & \textbf{78.89} \\ \midrule
ERM            & \multirow{3}{*}{AlexNet} & 48.44          & 48.44          & 48.43          \\ 
IRM            &                          & 70.67          & 71.16          & 70.03          \\  
\textbf{ISRL} &                          & \textbf{81.11} & \textbf{83.04} & \textbf{80.67} \\ \midrule
ERM            & \multirow{3}{*}{ResNet}  & 48.67          & 48.95          & 47.79          \\ 
IRM            &                          & 71.11          & 73.84          & 70.38          \\ 
\textbf{ISRL} &                          & \textbf{87.56} & \textbf{87.77} & \textbf{84.59} \\ \midrule
ERM            & \multirow{3}{*}{VGGNet}  & 47.56          & 47.35          & 47.27          \\ 
IRM            &                          & 68.89          & 72.83          & 68.56          \\ 
\textbf{ISRL} &                          & \textbf{79.79} & \textbf{80.96} & \textbf{79.77} \\ \midrule
\end{tabular}
\label{table:clf_SOTA_comp}
\end{table}

\begin{figure}[htbp]
    \centering
    \includegraphics[width=1.0\linewidth, trim = 0cm 0cm 0cm 0cm]{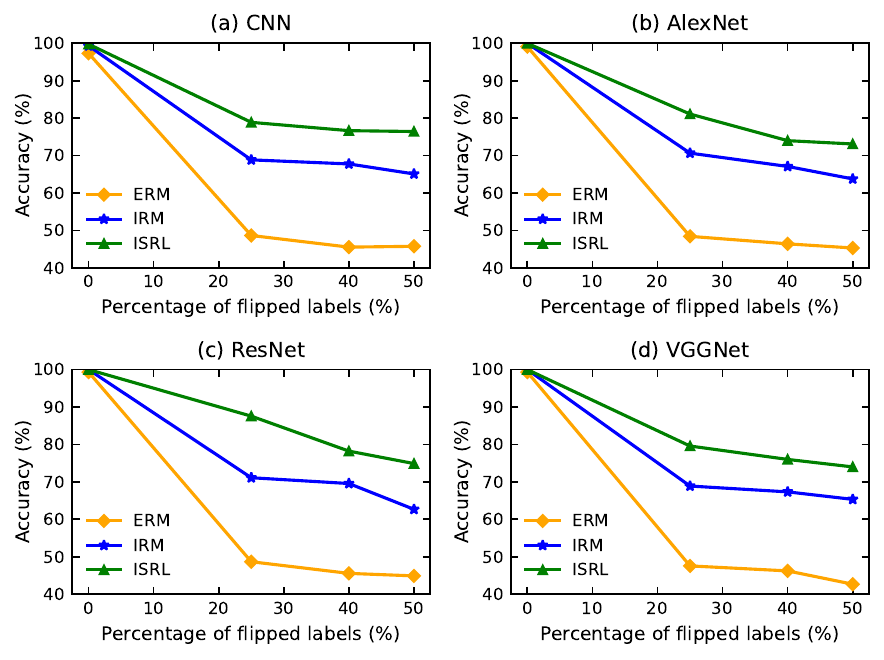} 
    \caption{A comparison of the baselines and ISRL on four different backbones over increased probability of label flipping. ISRL exhibits superior generalization across backbones under increasing label noise, suggesting enhanced invariance in its learned representations.}
    \label{fig:label_prob}
\end{figure}

\begin{table*}[htbp]
\centering
\caption{A comparison of classification accuracy on real 3D brain MRIs over all models with different network backbones.}
\begin{tabular}{lcccccccccccc}
\toprule
&\multicolumn{3}{c}{\textit{CNN}} & \multicolumn{3}{c}{\textit{AlexNet}} & \multicolumn{3}{c}{\textit{ResNet}} & \multicolumn{3}{c}{\textit{VGGNet}} \\
      \midrule
\textit{Model} & Acc.     & Prec. &  F1-sc.  & Acc.     & Prec. &  F1-sc.   & Acc.     & Prec. &  F1-sc.  & Acc.     & Prec. &  F1-sc.  \\
 \midrule
ERM  & 66.00  & 69.87 &  \multicolumn{1}{c|}{64.27}  & 62.00   & 62.10 &  \multicolumn{1}{c|}{61.92}   & 70.67   & 73.49 &  \multicolumn{1}{c|}{69.76}  & 62.67   & 72.80   & 58.00  \\
IRM   & 69.33   & 69.56 &  \multicolumn{1}{c|}{69.25}  & 68.00   & 72.00   &  \multicolumn{1}{c|}{66.47}   & 72.67   & 76.65 &  \multicolumn{1}{c|}{71.60}  & 65.33   & 65.89 &  65.03  \\
\midrule
CORAL   & 71.11   & 70.00 &  \multicolumn{1}{c|}{73.16}  & 70.00   &   {72.73}  &  \multicolumn{1}{c|}{71.54}   & 75.56   & 72.50 &  \multicolumn{1}{c|}{79.31}  & 76.67   & 82.73 & 81.54   \\
DANN   &  73.33  & 70.34 &  \multicolumn{1}{c|}{ {77.96}}  & 71.11    &  70.00  &  \multicolumn{1}{c|}{73.16}   & 75.19   & 71.67 & \multicolumn{1}{c|}{ {80.48}}  & 77.78   & 72.12 &  {73.87}   \\
IWDANN   & 73.33   & 74.29 &  \multicolumn{1}{c|}{72.07}  & 72.22   & 69.32   &  \multicolumn{1}{c|}{67.31}   & 75.93   & 72.50 &  \multicolumn{1}{c|}{79.31}  & 77.78   & 72.90 & 72.90   \\
CLoVE   & 74.00   & {74.04} &  \multicolumn{1}{c|}{73.98}  & 75.56    & 72.50   &  \multicolumn{1}{c|}{\textbf{77.80}}   &  {80.37}   &  \textbf{81.43}  &  \multicolumn{1}{c|}{78.97}  &  \textbf{82.22}  & 80.83 &\textbf{83.12}   \\
\textbf{ISRL} & \textbf{76.67}   &  \textbf{76.67} &  \multicolumn{1}{c|}{\textbf{76.66}}  & \textbf{76.00}   & \textbf{77.23}  &  \multicolumn{1}{c|}{75.72}   & \textbf{84.44}   & 80.37 &  \multicolumn{1}{c|}{\textbf{86.77}}  & \textbf{82.22}   & \textbf{82.73} &  81.91 \\
\bottomrule
\end{tabular}
\label{clf_brain}
\end{table*}

Fig.~\ref{fig:label_prob} illustrates the classification accuracy of all methods with an increased percentage of flipped labels (where $p^e=\{0, 0.25, 0.4, 0.5\}$) on four network backbones. All models demonstrate similar accuracy under ideal conditions of $0\%$ label flipping and no spurious correlations between the color information and image labels. However, as the percentage of label flipping increases, we observe a substantial performance drop. The baseline models experience a decrease of $30$-$55\%$ in accuracy vs. our proposed model shows a much lower drop of approximately $13$-$20\%$. Overall, our model consistently archives higher accuracy ($>10\%$) than the baselines. This indicates that ISRL is more robust when confronted with unseen data from a different distribution in the testing environment.

Tab.~\ref{clf_brain} reports a comparison of binary classification (AD vs. CN) results on 3D real brain MRIs for all backbones, as well as related domain generalization methods that focus on image intensity features (CORAL~\cite{sun2016deep}, DANN~\cite{ganin2016domain}, IWDAN~\cite{tachet2020domain}, and CLOvE~\cite{wald2021calibration}). Our ISRL model shows superior performance with consistently improved classification accuracy. Note that such a classification task on human brain images with large age variations is challenging, given that age has been demonstrated to induce anatomical brain changes~\cite{zhao2020training,rasal2022deep}. A typical classification network based on ERM achieves $\leq 70\%$ accuracy on our dataset. 

Similarly, Tab.~\ref{table:clf_cardiac_SOTA_comp} reports the classification performance on the real 3D cardiac MRI videos (classifying subjects with scar vs. non-scar). Our model ISRL consistently outperforms all methods, including IRM and domain generalization-based models. 

\begin{table}[htbp]
    \centering
    \caption{Accuracy comparison on real 3D cardiac MRI videos over all models under ViT and ViViT network backbones.}
    \begin{tabular}{lcccccc}
\toprule
& \multicolumn{2}{c}{\textit{ViViT}} & \multicolumn{2}{c}{\textit{ViT}}  \\
      \midrule
 \textit{Model} & Acc.        & F1-sc.  & Acc.        & F1-sc.     \\
 \midrule
ERM   & 79.67  &  \multicolumn{1}{c|}{79.77}  & 78.65   &  77.93      \\
IRM   &  {86.51}   &  \multicolumn{1}{c|}{ {86.43}}  & 82.02   &  81.96   \\
\midrule
CORAL   & 83.15   &  \multicolumn{1}{c|}{83.87}  & 79.78  & 76.32    \\
DANN   & 80.90   &  \multicolumn{1}{c|}{80.00}  & 79.78   & 81.63    \\
IWDANN   & 83.15   &  \multicolumn{1}{c|}{82.76}  & 80.90  & 80.46    \\
CLoVE   & 84.24   &  \multicolumn{1}{c|}{83.67}  &  {85.39}  &  {85.06}    \\
\textbf{ISRL}   & \textbf{88.76}   &  \multicolumn{1}{c|}{\textbf{88.76}}  & \textbf{87.64}   &  \textbf{87.63}  \\
\bottomrule
\end{tabular}

\label{table:clf_cardiac_SOTA_comp}
\end{table}

\begin{figure}[htbp]
    \centering
    \includegraphics[width=1.0\linewidth, trim = 0cm -0.5cm 0cm 0cm]{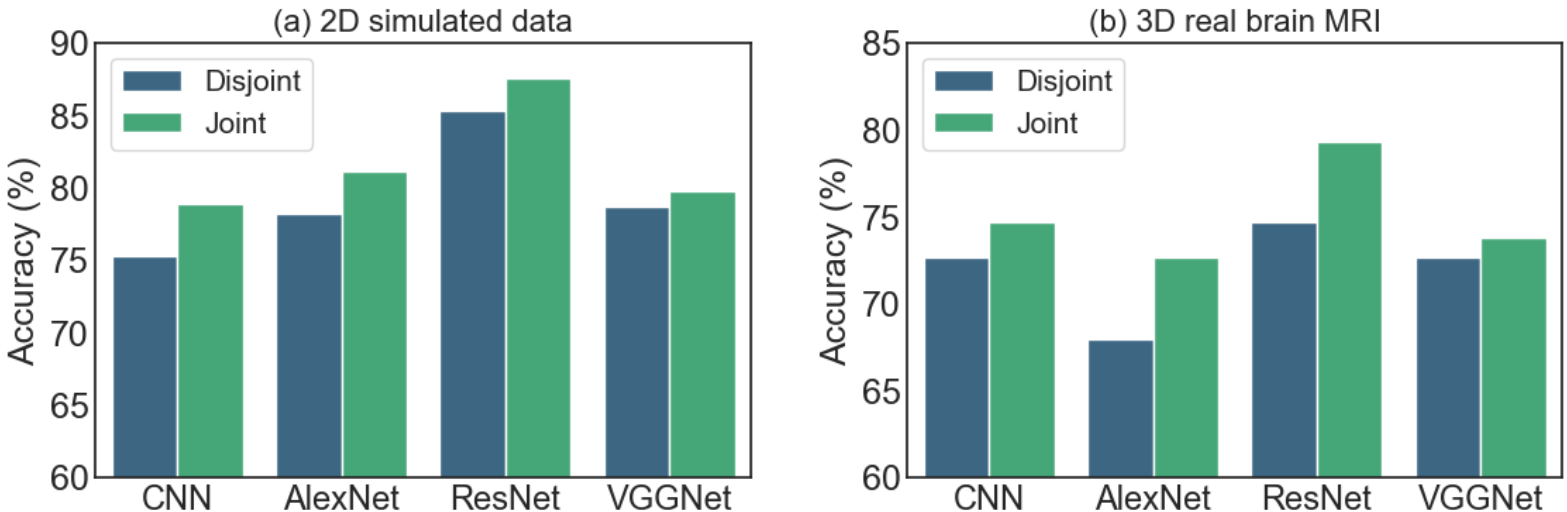} \caption{Comparison of disjoint (two-step) learning vs. our joint approach on both 2D simulated data and 3D brain MRIs on different network backbones. Joint optimization of shape learning and invariant classification yields superior performance.}
    \label{fig:opt_comp}
\end{figure}

Fig.~\ref{fig:opt_comp} reports the performance comparison of our joint learning model with a disjoint (two-step) training approach on both 2D simulated and 3D brain datasets (based on all network backbones). Our joint learning model consistently outperforms the two-step approach by approximately $2$-$3\%$.

\captionsetup[figure]{belowskip=0pt}

Tab.~\ref{penalty_2D_3D_brain} displays the classification performance of varying penalty weights ($\lambda$) for both IRM and our model ISRL on different backbones. It shows that a small value of $\lambda$ weakens the regularizer's effect and reduces the classifier's efforts to search for invariant representations, while a larger value of $\lambda$ can result in overly emphasized commonalities at the cost of domain-specific performance. As $\lambda$ increases, both models' performance declines beyond a threshold. 

\begin{table}[htbp]
    \centering
    \caption{Classification accuracy (\%) comparison across various penalty weights ($\lambda$) under ResNet backbone.}
    \begin{tabular}{lcccc}
\toprule
  \textit{Dataset}  & \multicolumn{2}{c}{\textit{2D simulated data}} & \multicolumn{2}{c}{\textit{3D brain MRI}} \\
    \midrule
Penalty  & IRM               & ISRL             & IRM             & ISRL          \\
\midrule
10000          & 62.67             & 74.89             & 65.33           & 74.67          \\
25000          & 65.11             & 76.44             & 66.67           & 79.33          \\
50000        & 69.56             & 85.33             & 72.67           & \textbf{84.44}          \\
75000          & 71.11             & \textbf{87.56}            & 68.00           & 76.00          \\
100000         & 68.67             & 78.72             & 68.67           & 78.00   \\
\bottomrule
\end{tabular}
    \label{penalty_2D_3D_brain}
\end{table}

Tab.~\ref{tab:feat_ab} presents an ablation study evaluating the effectiveness of geometric shape features, $\mathcal{G}_E$, and invariant learning mechanisms. We first compare the classification model in an empirical learning setting without invariant learning,  with or without geometric features. We then compare with the introduced invariant learning mechanism. While training with geometric shape features improves classifier performance over using only image features, $\mathcal{I}_E$, our ISRL (integrating invariant learning of image and shape features) achieves the best performance, surpassing the IRM that operates solely in the image space.

\begin{table}[htbp]
    \centering
    \caption{An ablation study on the effectiveness capturing invariant geometric shape features, $\mathcal{G}_E$, with image features, $\mathcal{I}_E$, on both simulated 2D and real 3D datasets.}
    \begin{tabular}{lcccc}
\toprule
\multirow{2}{*}{Datasets} & \multicolumn{2}{c}{Non-Invariant} & \multicolumn{2}{c}{Invariant} \\
& $\mathcal{I}_E$              & $\mathcal{I}_E + \mathcal{G}_E$  & $\mathcal{I}_E$         & \textit{ISRL}     \\
                         \midrule
 2D Quick-draw                     & $48.67$           & \multicolumn{1}{c|}{$52.22$}    & $71.11$       & $\textbf{87.56}$     \\
3D Brain                    & $70.67$           & \multicolumn{1}{c|}{$74.00$}   & $72.67$        & $\textbf{84.44}$     \\
3D Cardiac                  & $79.67$           & \multicolumn{1}{c|}{$85.39$}  & $86.51$         & $\textbf{88.76}$   \\
                         \bottomrule
\end{tabular}
    \label{tab:feat_ab}
\end{table}
\raggedbottom

\begin{figure}[t]
    \centering
    \includegraphics[width=0.93\linewidth, trim = 0cm 0cm 0cm 0cm]{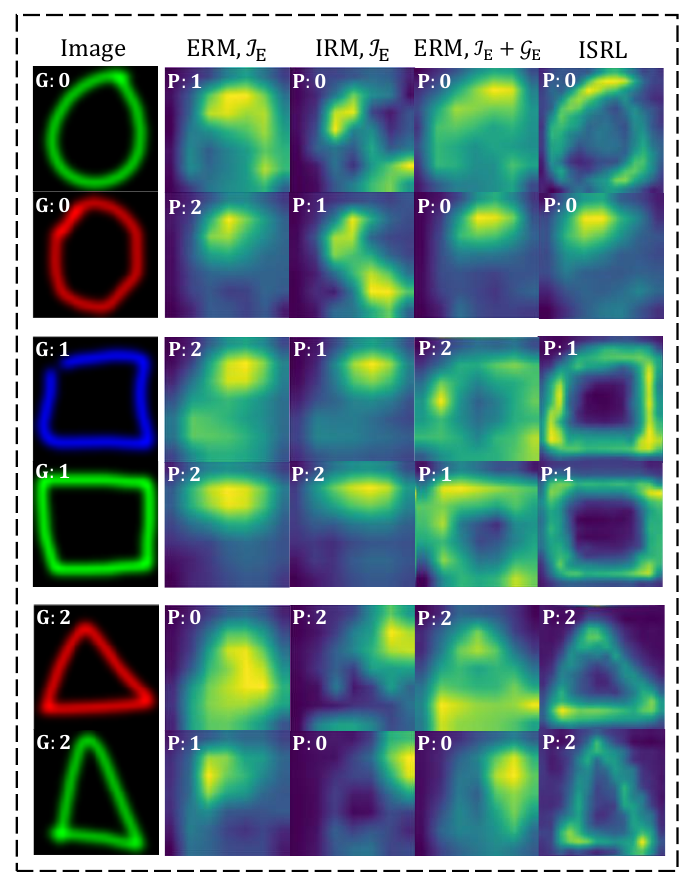} 
    \caption{A visual comparison of activation maps generated by ERMs and invariant models (IRMs, including ISRL). G: Ground-Truth, P: Prediction.}
    \label{fig:feat_viz}
\end{figure}

Fig.~\ref{fig:feat_viz} visualizes activation maps~\cite{zeiler2014visualizing} of the last convolution layers for various baseline models under ResNet backbones. It shows that ERM, when incorporating shape features alongside image features, captures more relevant object-focused representations than when using image features alone. However, our ISRL model, utilizing an invariant feature learning mechanism in integrated shape and image spaces, further improves the network attention to object shapes across all classes, demonstrating its enhanced capacity to capture relevant geometric structures effectively. \\

{\noindent \bf Limitation \& Discussion.} Intuitively, IRM captures factors that have invariant relations with the label to make robust predictions in different environments. The connection between IRM and causality originates from the invariance of causal mechanism in causal inference theory, where the relation between each variable and its ``parent'' variables is invariant. Such a connection was discussed in the original IRM paper and follow-up works \cite{arjovsky2019invariant,chang2020invariant,lu2021invariant,wang2022out}. Existing studies have shown that IRM effectively removes spurious correlations under different causal models with certain assumptions~\cite{liu2021federated,zare2022removal,lin2022bayesian}. However, if the real-world data does not meet such assumptions (e.g., when no causal variables are available for the label, or the spurious correlation is stronger than the invariant relations in all environments), IRM may not outperform other methods. Additionally, IRMs can result in suboptimal predictors when training distributions fail to sufficiently cover the full range of potential test scenarios, or when unobserved confounders affect both input features and target variables across environments;  leading to limited generalization performance. As a result, the model may struggle to adapt to new data distributions, thereby weakening its robustness and reliability in real-world applications. Addressing the limitation of IRM is out of the scope of this paper, our focus is harnessing its effectiveness in suitable scenarios.
\captionsetup[figure]{belowskip=0pt}

While our current experimental design on the real-world data is in the context of a specified environment, it is noteworthy that our proposed ISRL can be naturally general to multi-environment settings. Future work will explore other potential variables, such as sex and genetics~\cite{rasal2022deep} or batch effects due to imaging sites~\cite{wachinger2021detect,wen2020convolutional}.

\section{Conclusion}
This paper introduces a novel model, named ISRL, that for the first time develops invariant shape representation learning for image classification. In contrast to previous approaches that derive features in image space, our model jointly learns the invariant features to image labels in an integrated space of images and geometric deformations. To achieve this, we develop a new learning paradigm that captures the invariant shape and image features across multiple training environments by minimizing the risk of spurious correlations that might be present in the shifted data distributions. Experimental results show that our method significantly improves the classification performance on simulated 2D data, real 3D brain images, and cardiac MRI videos. 

Our work on ISRL is an initial attempt to explore the power of learning causal shape representations for image analysis tasks. Future directions include but not limited to 
i) investigating the interpretability of ISRL in shape-based deep networks, ii) generalizing the model's ability to learn causally invariant image and shape features from multimodal image data, and iii) developing a robust classification model to deal with unknown hidden spurious factors. Besides, investigating the problem under more issues in practice (such as insufficient labels, noises, selection biases, and privacy concerns) would also be an interesting direction.

\section*{Acknowledgements}
\noindent This work was supported by NSF CAREER Grant 2239977 and NIH 1R21EB032597.

{\small
\bibliographystyle{ieee_fullname}
\bibliography{egbib}
}

\end{document}